\documentclass[dvipsnames, 11pt,a4paper]{article}
\usepackage[hyperref]{acl2020}
\usepackage{times}
\usepackage{mwe}
\usepackage{latexsym}
\usepackage{multirow}
\usepackage{graphicx}
\usepackage{lipsum}
\usepackage{xcolor}
\definecolor{drawiored}{RGB}{234,107,102}

\def\B{{\bf B}}

\def\I{{\bf I}}
\def\T{{\bf T}}

\def\0{{\bf 0}}
\def\1{{\bf 1}}

\usepackage{booktabs}
\usepackage{microtype}

\aclfinalcopy 
\def\aclpaperid{1322} 

\title{To Find Waldo You Need Contextual Cues: Debiasing \textit{Who's Waldo}}

\author{Yiran Luo \quad Pratyay Banerjee \quad Tejas Gokhale \quad Yezhou Yang \quad Chitta Baral\\

  Arizona State University, Tempe, AZ, USA \\
  \texttt{\{yluo97, pbanerj6, tgokhale, yz.yang, chitta\}@asu.edu} \\
  
  }

\date{}

\begin{document}

\maketitle
\begin{abstract}
We present a debiased dataset for the Person-centric Visual Grounding (PCVG) task first proposed by~\citet{Cui_2021_ICCV} in the \textit{Who's Waldo} dataset.
Given an image and a caption, PCVG requires pairing up a person's name mentioned in a caption with a 
bounding box that 
points to the person in the image.
We find that the original \textit{Who's Waldo} dataset compiled for this task contains a large number of biased samples that are solvable simply by heuristic methods; for instance, in many cases the first name in the sentence corresponds to the largest bounding box, or the sequence of names in the sentence corresponds to an exact left-to-right order in the image.
Naturally, models trained on these biased data lead to over-estimation of performance on the benchmark.
To enforce models being correct for the correct reasons, we design automated tools to filter and debias the original dataset 
by ruling out all examples of insufficient context, such as those with no verb or with a long chain of conjunct names in their captions. 
Our experiments show that our new sub-sampled dataset\footnote{Available at: \url{https://github.com/fpsluozi/tofindwaldo}} contains less bias with much lowered heuristic performances and widened gaps between heuristic and supervised methods. 
We also demonstrate the same benchmark model trained on our debiased training set outperforms that trained on the original biased (and larger) training set on our debiased test set. 
We argue our debiased dataset offers the PCVG task a more practical baseline for reliable benchmarking and future improvements. 
\end{abstract}

\section{Introduction}



\begin{figure}[t]
    \centering
    \includegraphics[width=\linewidth]{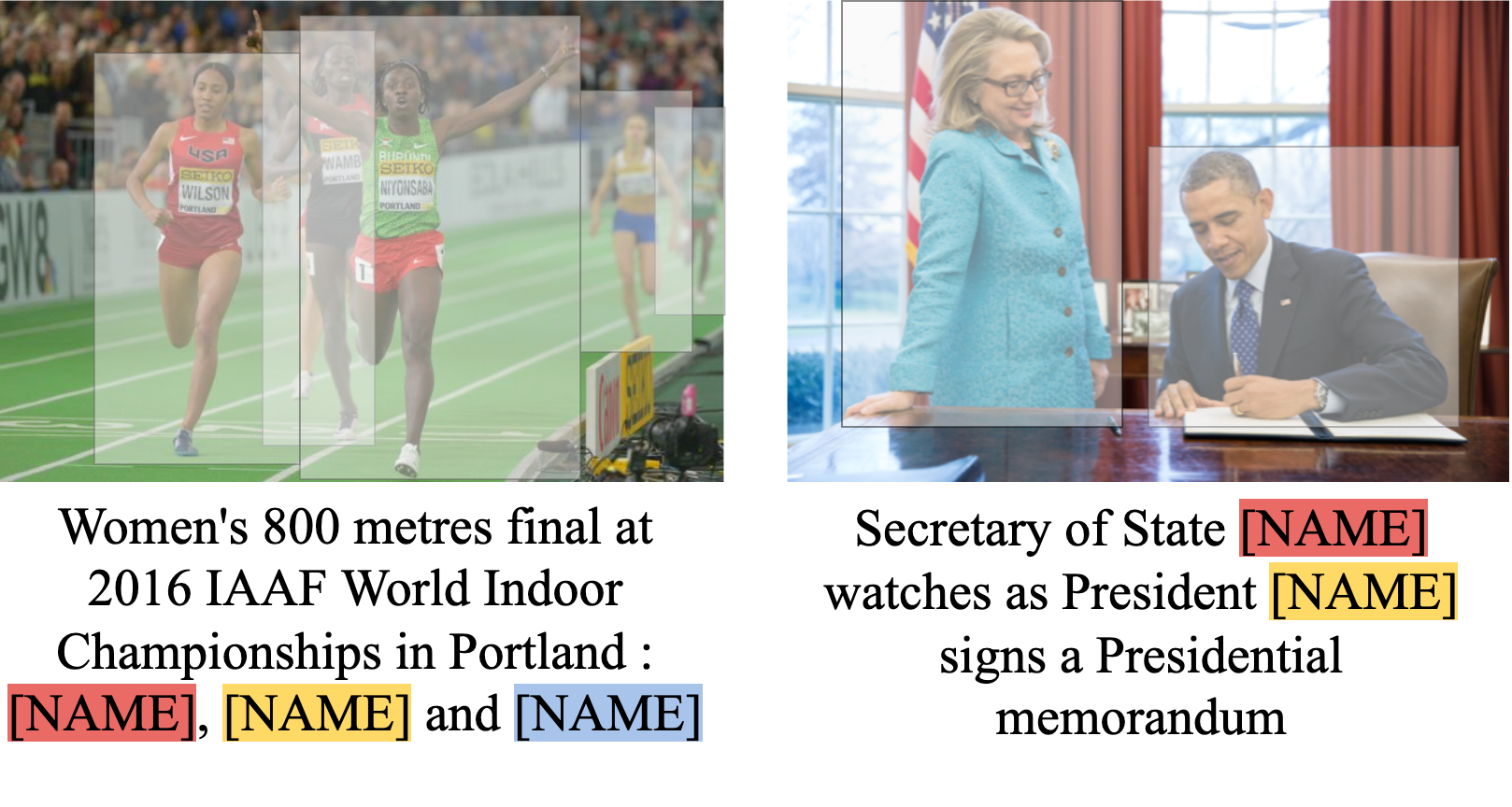}
    \caption{
    We find many biased data from the original Who's Waldo dataset contain insufficient contextual cues and cannot be used to map names to persons in an image.
    \textbf{Left:} An unsolvable example with no actions nor descriptions w.r.t the detected persons. 
    Given no background knowledge about the individuals, one can only guess the masked \texttt{[NAME]}'s based on heuristic biases such as the locations of the bounding boxes.  \textbf{Right:} A qualifying example with clearly worded interactions (e.g. detectable verbs such as 'watches' \& 'signs') about each masked name - the very type of data we incorporate into our debiased dataset. }
    \label{figure1}
\end{figure}

A newly released task called Person-centric Visual Grounding \cite{Cui_2021_ICCV} poses an interesting angle into contextual reasoning in vision-language. 
The task is motivated by humans' reasoning ability. 
Humans viewing an image with a caption as shown in Figure~\ref{figure1} can reason (and if needed, speculate) which name refers to which person in the image.
This reasoning task involves multiple abilities, such as perceiving characteristics and behaviors of people, understanding their actions in context, speculating about their intentions and effects human of actions~\cite{fang2020video2commonsense}, and connecting visually perceived characteristics with grounded descriptions in natural language~\cite{kazemzadeh2014referitgame,yu2016modeling,vcr}.
In many cases, this task can be performed without knowing the names of the people; for instance in the example on the right, one person is signing and the other is not, as such it is possible to predict which person refers to President and Secretary of State respectively.
However, in cases such as the example on the left, if all persons are performing the same action (running on a track), then it is hard to match names with these runners without any additional information.
Progress in the PCVG task can thus help better capture what exact contextual cues are needed to learn about a person's characteristics in a scenario, and can aid improvements in visual understanding about human interactions and behaviors.

To support this task, ~\citet{Cui_2021_ICCV} offer a large-scale dataset called \textit{Who's Waldo} which consists of 272K annotated real life images. 
Ideally, the dataset should consist of input-output pairs (such as the example on the right in Figure~\ref{figure1}) which are `solvable' as opposed to the one on the left which is ambiguous.
However, as we explore the original \textit{Who's Waldo} dataset, we encounter a great portion of cases that resemble the left example in Figure \ref{figure1}, unsolvable data with insufficient contextual cues. 
Given such context, if we do not recognize who exactly is in the picture, even we human beings cannot tell which name is who. 
We can then only make predictions with biased assumptions, such as the first named person would always be on the leftmost, or the main subject would always make up the largest area. 
Such biases in the original dataset may explain why the heuristic methods perform very strongly, outperforming random guessing by a big $27\%$ increase in test accuracy and trailing the top benchmark only by $6\%$. 
We believe a fair dataset should not encourage approaches to adopt biases to such an extent, and thus the original baseline model overestimates its performance. 

Inspired by dataset debiasing works such as VQA-CP \cite{vqa-cp} and GQA-OOD~\cite{kervadec2021roses}, we create a debiased collection of 84K annotated image-captions out of the \textit{Who's Waldo} dataset by filtering out all biased data with insufficient context. 
We evaluate the quality of our new dataset by applying the original heuristic methods as well as \textit{Who's Waldo}'s benchmark model. 
Results show that our debiased dataset greatly reduces the heuristic biases from the original dataset and provides the PCVG task a more practical baseline for future developments.

\section{Related Work}
\textbf{Dataset Debiasing.} 
We take many inspirations from previous studies on uncurated datasets. 
A task dataset if not curated properly could lead to methods that cheat their ways through without learning generalized information. For example, VQAv2 \cite{vqav2} addresses the imbalance between language and images in VQAv1 \cite{vqav1} which results in visual information being ignored and inflated model performance.
VQA-CP~\citep{vqa-cp} and GQA-OOD~\citep{kervadec2021roses} were designed to test model performance if spurious correlations exist in the training dataset.
\citet{cadene2019rubi,chen2020counterfactual,gokhale2020mutant} are bias-aware techniques that mitigate dataset bias with modeling and data augmentation.
\citet{vcr-shortcut} introduce exploits by matching repeated texts in questions and answers to achieve high scores in Visual Commonsense Reasoning \cite{vcr}.

We also learn from various techniques to amend priors, biases, or shortcuts in datasets. REPAIR \cite{repair} uses resampling to fix representation biases in image datasets. ~\citet{dasgupta-sent-embedding} incorporate compositional information into sentence embeddings for Natural Language Inference. DQI \cite{dqi} offers quantitative metrics to assess biases in automated dataset creation in Natural Language Processing. ~\citet{adv-filters} introduce adversarial measures to mitigate biases in various Natural Language Processing and Computer Vision tasks.    

\noindent\textbf{Visual Grounding.} The PCVG task adapts previous supervised Visual Grounding models as its original baselines. The Visual Grounding task is defined as locating specific objects in an image from a textual description. First established by \citet{karpathy2014deep}, following researches have evolved into extracting attention information such as works by \citet{deng2018visual} and \citet{endo2017attention}. A huge variation of datasets for Visual Grounding have also been created, including Flicker30k \cite{plummer2015flickr30k}, Visual Genome \cite{krishna2017visual}, and RefCOCO \cite{yu2016modeling}.

\noindent \textbf{Referring Expression Comprehension (REC).} An active branch from Visual Grounding, the Referring Expression Comprehension task \cite{rohrbach2016grounding} is no longer restricted to object categories. Instead its goal is to relate a free region in an image to a sentence description. Mattnet \cite{yu2018mattnet} is one prominent approach that leverages both attention features and relation extraction for the objects in the image. \citet{qiao2020referring} offers a comprehensive survey on this topic.

\noindent \textbf{Human Detection.} A specialized category under Object Detection, detecting humans with bounding boxes in images nowadays can easily use open source toolboxes including MMDetection \cite{mmdetection} or Detectron \cite{wu2019detectron2} that are trained on large-scale real life image datasets like COCO \cite{lin2014microsoft}. Recent works such as DarkPose \cite{zhang2020distribution} also attempt to utilize human pose information to better single out human traits from complex background. 



\section{Method}

\begin{table}

    \centering
    \small
    \resizebox{\linewidth}{!}{
    \begin{tabular}{@{}lrrrr@{}}
    \toprule 
    \textbf{Selection} & \textbf{Train} & \textbf{Val.} & \textbf{Test} & \textbf{Unused}\\
    \midrule
    Original & 179073 & 6740 & 6741 & 79193 \\
    \texttt{no-verb} & 125585 & 3446 & 3529 & 34366 \\
    \texttt{conjunct-names} & 16446 & 2237 & 2227 & 15693 \\
    \midrule
    Ours & 45884 & 2102 & 2049 & 33611 \\
    \bottomrule
    \end{tabular}
    }
    
    \caption{\label{stat-table} The data in our debiased dataset are filtered and regrouped from all four splits in the original. Notice, examples such as the \textbf{Left} in Figure \ref{figure1} can have both zero verb and at least three conjunct names.}
\end{table}

\begin{figure*}[h]
    \centering
    \includegraphics[width=\linewidth]{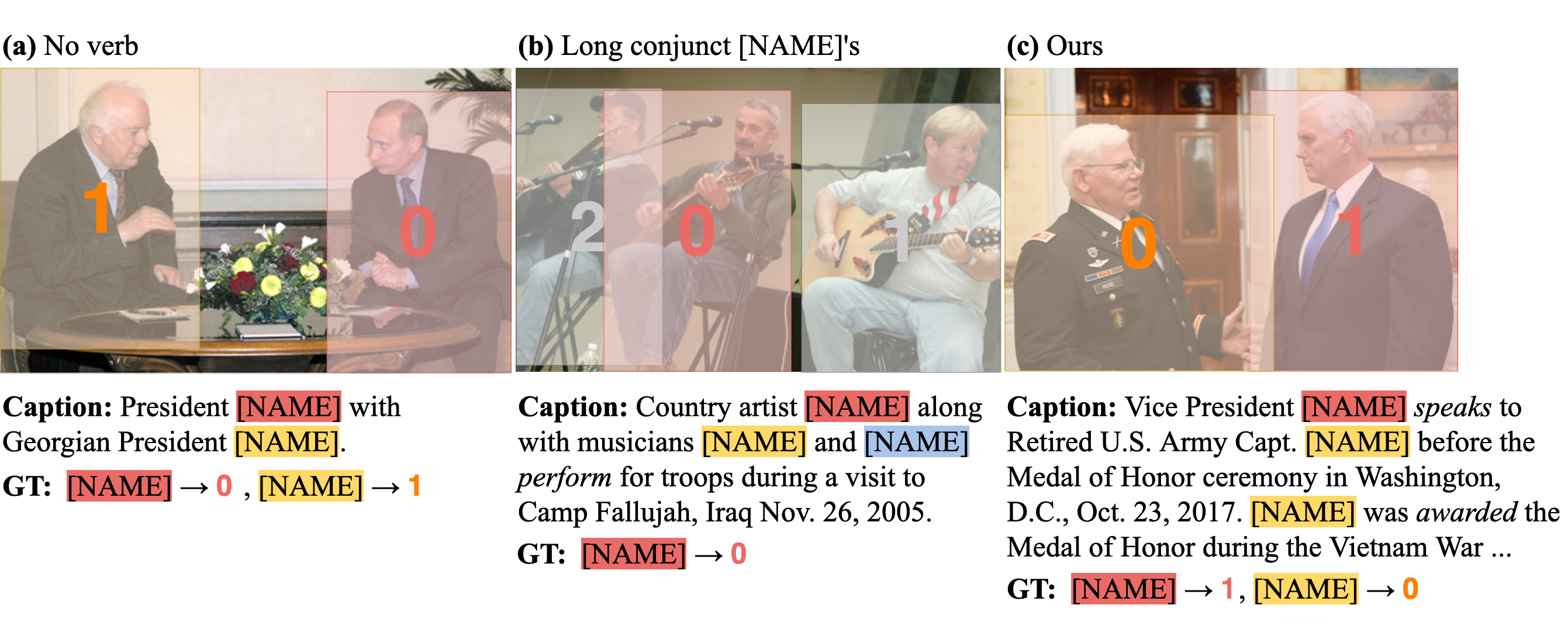}
    \caption{
        \textbf{(a)} and \textbf{(b)} represent the two major types of insufficient and biased data that we filter out. 
        \textbf{(c)} represents the ones we choose for our debiased dataset. 
        We label all detected verbs in \textit{italic}. 
        We apply color coding to indicate different person entities in a caption. 
        We also use gray bounding boxes to refer to those 'incorrect options' not included in ground truth, such that in the ground truth of \textbf{(b)}, the only pair we need to associate is \colorbox{drawiored}{[NAME]} with Bounding Box 0, while the two other bounding boxes serve as mere distractions. 
        }
    \label{figure2}
\end{figure*}

In this section, we introduce the Person-centric Visual Grounding task, discuss the original \textit{Who's Waldo} dataset, and provide our analysis of shortcuts, biases, and other issues that we discovered in the dataset.
We describe the process via which we curate, debias, and filter the dataset.

\begin{table*}[h]
    \centering
    \footnotesize
     \resizebox{.9\textwidth}{!}{
\begin{tabular}{lllrrr} 
\toprule
\textbf{Method}             & \textbf{Training Set} & \textbf{Test Set}                                                          & \textbf{Test Accuracy} & \textbf{$\Delta_r$} & \textbf{$\Delta_h$}  \\ 
\hline
Random                      & –                     & Original~Test                                                              & 30.9                 & 0.0                 & –                    \\
Big $\rightarrow$ Small     & –                     & Original~Test                                                              & 48.2                 & +17.3               & –                    \\
L $\rightarrow$ R (All)     & –                     & Original~Test                                                              & 38.4                 & +7.5                & –                    \\
L $\rightarrow$ R (Largest) & –                     & Original~Test                                                              & 57.7                 & +26.8               & 0.0                  \\
Gupta et al.                      & COCO              & Original~Test                                                            & 39.3                 & +8.4                & -18.4                \\
SL-CCRF                     & Flickr30K Entities   & Original~Test                                                              & 46.4                 & +15.9               & -11.3                \\
MAttNet                     & RefCOCOg              & Original~Test                                                              & 44.0                 & +13.1               & -13.7                \\

\textit{Who's Waldo}        & Original Train        & Original~Test                                                              & 63.5                 & +32.6               & +5.8                 \\ 
\hline
Random                      & –                     & Our Test                                                                   & 31.0                 & 0.0                 & –                    \\
Big $\rightarrow$ Small     & –                     & Our Test                                                                   & 43.8                 & +12.8               & –                    \\
L $\rightarrow$ R (All)     & –                     & Our Test                                                                   & 32.4                 & +1.4                & –                    \\
L $\rightarrow$ R (Largest) & –                     & Our Test                                                                   & 44.3                 & \textbf{+13.3}      & 0.0                  \\
\textit{Who's Waldo}        & Original Train        & Our Test                                                                   & 50.2                 & +19.2               & +5.9                 \\

\textit{Who's Waldo}        & Our Train             & Our Test                                                                   & \textbf{54.0}        & +23.0               & \textbf{+9.7}        \\
\hline
\textit{Who's Waldo}        & Our Train             & \begin{tabular}[c]{@{}l@{}}\textit{Biased samples} of\\Original Test\end{tabular} & 48.2                 & –                   & –                  \\
\bottomrule
\end{tabular}
     }
    \caption{\label{tableacc} Evaluation on the test sets using the original \textit{What's Waldo} and our debiased dataset. $\Delta_r$ denotes relative improvement over random guessing, and $\Delta_h$ denotes relative improvement over the best heuristic. The \textit{biased samples} represents a total of 4.7K samples from the original test set that are filtered out by our debiasing procedure. The original work also compares its baseline performance with multiple pre-trained visual grounding models, such as~\citet{gupta2020contrastive} trained with COCO~\cite{lin2014microsoft}, SL-CCRF~\cite{slccrf} trained with Flickr30K Entities~\cite{plummer2015flickr30k}, and MAttNet~\cite{yu2018mattnet} trained with RefCOCOg~\cite{mao2016refcocog}. All reported accuracies in this table are the strongest averaged performances per setting and fall within a fluctuation of $\pm 1\%$ .}
\end{table*}

\subsection{The Task}
The Person-centric Visual Grounding task is defined as follows. The givens are an image \I, a set of m $\geq$ 1 person detections \B $ $ (in form of bounding boxes), and a corresponding image caption \T $ $  where its tokens contain references to n $\geq$ 1 persons. For each referred person, we look for the best matching detection from the givens. We also assume no two persons can be matched with the same detection.

\subsection{The \textit{Who's Waldo} Dataset}
The dataset consists of 272K real-life captioned images sourced from the free Wikimedia Commons repository. 
Each image pictures individuals under the 'People by name' category on Wikimedia Commons, while its caption describes the scene and explicitly mentions the featured people in real names.
Key dataset creation procedures, text pre-processing, identifying person entities in captions, detecting bounding boxes of people in images, and generating ground truths linking bounding boxes and names, are all done with existing automated tools such as FLAIR \cite{flair} and MMDetection \cite{mmdetection}. 
To prevent misuse, in the publicly released version, all the real names in the captions are replaced with the \texttt{[NAME]} token, but references between bounding boxes and token indices are given in individual annotation files. 
This is equivalent to masking each name with indexed placeholders such as \texttt{PERSON1}, \texttt{PERSON2}, etc.
Amongst the entirety of 272K annotated samples, 179K samples are used for training, 6.7K for validation, and 6.7K for testing.
Each test sample is supposed to either \textit{mention at least two persons} or \textit{choose from at least two bounding boxes}. 
The original test set is further validated manually on Amazon Mechanical Turk. 

\subsection{Biases in \textit{Who's Waldo}}

The premise of the Person-centric Visual Grounding task is to use ONLY the caption text and the image as the cues to find out the correct bounding box from the image per mentioned name. However, we observe a large portion of the original \textit{Who's Waldo} dataset does not provide sufficient contexts and can only be solved by heuristic methods. We discuss two major types of biases that we discover in the following sections.

The first type \texttt{no-verb} is that the caption text contains zero detectable verbs. Since linguistically a verb is the crucial part of an action that assigns participants with semantic roles, we technically have no way to tell who performs or who receives an action without verbs. For example in Figure \ref{figure2}(a), we are unable to tell who is who from the image and the no-verb caption alone, unless we recognize Vladimir Putin or the Georgian President with external knowledge. 

The second type \texttt{conjunct-names} is that the caption contains a long chain of conjunct referred names. Shown in Figure \ref{figure2}(b), all the referred names share the verb \textit{perform}, joined together only with conjunct words such as \textit{and} or \textit{along with}. With no indication of the order amongst these persons, we can only resort to a naive positional order such as left-to-right. But since we may also have extra bounding boxes as choices, such naive assumption is indeed unreliable. Figure \ref{figure2}(b) is such an example that the first mentioned name is not always the one in the left-most bounding box.

\subsection{Data Curation for De-biasing}

In order to resolve the aforementioned limitations of the original dataset, we utilize two pipelines in SpaCy ver 3.0 \cite{spacy} to filter out the biased data. We apply the POS-Tagging pipeline to find out if sentences in an image caption contain verbs in any form of conjugation. In parallel, we use the Dependency Parsing pipeline to examine if any \texttt{[NAME]} token conjuncts with more than one \texttt{[NAME]}'s from different referred persons. We jointly filter out any example that either (a) contains zero verbs, or (b) has at least three conjunct referred person names in a sentence. For both pipelines, we replace the \texttt{[NAME]} tokens that refer to the same person in a caption with a random popular first name, so that the natural language-based SpaCy pipelines can yield more accurate results. Both pipelines use the state-of-the-art \texttt{en-web-core-trf} model which is built on RoBERTa \cite{roberta}. 

Ultimately, our filtering procedure produces 84K qualifying image-caption pairs. Table \ref{stat-table} shows the distribution of samples sourced from each split of the original through our two debiasing pipelines. We utilize data from the unused yet legitimately annotated 79K samples of the original dataset. We reorganize and split all the qualifying 84K samples into 74K for training, 5K for validation, and 5K for test. Our new test set does not overlap with the original training set. Similarly to the design of the original, we enforce that all samples in our new test set involves no trivial case that contains exactly one referred name and exactly one bounding box. We also make sure that any test set sample always has at least one name-to-bounding-box pair as ground truth. 

\section{Experiments and Baselines}
\noindent \textbf{Setup.}
We evaluate the quality of our debiased dataset with the same heuristic and Transformer-based methods from the original paper.
We also train the benchmark model on both the original and our new training set. 
We report the accuracies obtained from our new test set as the new baselines. 

\noindent \textbf{Heuristics.} We inherit the original heuristic measures to study the potential biases of our debiased dataset versus those of the original dataset. Alongside Random guessing, we assign the names in the caption to the bounding boxes sorted by: (a) decreasing area size (Big $\rightarrow$ Small), (b) left-to-right upper-left coordinates (L $\rightarrow$ R (All)), and (c) left-to-right upper-left coordinates of the largest $d$ bounding boxes, $d$ being the larger between the number of bounding boxes and the number of names in a test case (L $\rightarrow$ R (Largest)). 

\noindent \textbf{Transformer-based Models.} We adapt the original benchmark \textit{Who's Waldo} model to our debiased dataset and see how well it can perform under the updated contexts. The benchmark model is a multi-layer multi-modal Transformer \cite{transformer}. Based on UNITER \cite{uniter}, it learns to maximize the similarities between the corresponding person names and bounding boxes while minimize the similarities between those that do not match up. We fine-tune the \textit{Who's Waldo} model with pre-trained weights from UNITER.


\noindent \textbf{Analysis of Results.} Table \ref{tableacc} shows the test set accuracies for the original dataset and our debiased dataset. 
We find that the heuristic measures have overall lower performance on our new dataset, meaning we have successfully reduced the effects of the positional and the size-based biases from the original dataset. Most significantly, we have lowered L $\rightarrow$ R (All) from +7.5\% to +1.4\%, almost equal to randomness. Even the strongest L $\rightarrow$ R (Largest) heuristic has been lowered from +26.8\% all the way down to +13.3\% as well. Our dataset is thus proven less biased compared to the original. 

We also show that our dataset has better practicality for the task. Measured with our new test set, the performance of the \textit{Who's Waldo} benchmark model trained with the original training set performs 3.8\% lower than that trained with our new, smaller training set. Meanwhile, the test accuracy gap between the Transformer-based method and the heuristic methods has become larger using our debiased dataset, widened from 5.8\% to 9.7\%. In addition, using the filtered \textit{biased samples} from the original test set on our new trained model yields an even lower performance at 48.2\%, which indicates our new baseline model now adopts fewer biases during training compared to the original.  Altogether with the lowered new baseline accuracy of 54.0\%, we argue that our debiased dataset improves the quality of contextual cues that supervised models can learn from, and leaves more applicable room for improvements in the future.

\section{Conclusion}
We present a refined dataset for the PCVG task with samples that contain contextual information required for the task. 
We address prominent biases that we identified in the original task dataset by filtering out a large number of unsolvable cases, and report new baseline performances on the new benchmark.
Our refined dataset can serve as a more reliable benchmark to enable fair comparisons for new modeling techniques and training protocols.

\section*{Acknowledgements}
This research was supported by grants from DARPA SAIL-ON, DARPA KAIROS, NSF 1816039, and NSF 2132724.
The views and opinions of the authors expressed herein do not necessarily state or reflect those of the funding agencies and employers.

\section*{Ethical Considerations}
Our curated dataset is available at \url{https://github.com/fpsluozi/tofindwaldo} . We will also follow the same licensing and data sharing policy as the original Who's Waldo dataset.
\bibliography{anthology,acl2020}
\bibliographystyle{acl_natbib}


\end{document}